\documentclass[11pt,a4paper]{article}
\usepackage{diagbox}
\usepackage{authblk}
\usepackage[hyperref]{emnlp2020-edited-for-arXiv}
\usepackage{times}
\usepackage{latexsym}

\usepackage{microtype}

\usepackage{multirow}

\usepackage[T1]{fontenc}
\usepackage[acronym,shortcuts,smallcaps]{glossaries}

\newglossaryentry{LSTM}
{
  name={LSTM},
  description={long short-term memory},
  first={\glsentrydesc{LSTM} (\glsentrytext{LSTM})},
  plural={LSTMs},
  descriptionplural={long short-term memories},
  firstplural={\glsentrydescplural{LSTM} (\glsentryplural{LSTM})}
} 

\newglossaryentry{RNN}
{
  name={RNN},
  description={recurrent neural network},
  first={\glsentrydesc{RNN} (\glsentrytext{RNN})},
  plural={RNNs},
  descriptionplural={recurrent neural networks},
  firstplural={\glsentrydescplural{RNN} (\glsentryplural{RNN})}
} 

\newglossaryentry{NLP}
{
  name={NLP},
  description={natural language processing},
  first={\glsentrydesc{NLP} (\glsentrytext{NLP})},
} 

\newglossaryentry{MT}
{
  name={MT},
  description={machine translation},
  first={\glsentrydesc{MT} (\glsentrytext{MT})},
}

\newglossaryentry{ML}
{
  name={ML},
  description={machine learning},
  first={\glsentrydesc{ML} (\glsentrytext{ML})},
} 

\newglossaryentry{HAN}
{
  name={HAN},
  description={hierarchical attention network},
  plural={HANs},
  descriptionplural={hierarchical attention networks},
  first={\glsentrydesc{HAN} (\glsentrytext{HAN})},
  firstplural={hierarchical attention networks (\glsentryplural{HAN})}
} 

\newglossaryentry{HAN-ST}
{
  name={HAN$_{\textrm{ST}}$},
  description={hierarchical attention network with structure tags},
  descriptionplural={hierarchical attention networks with structure tags},
  first={\glsentrydesc{HAN-ST} (\glsentrytext{HAN-ST})},
  firstplural={hierarchical attention networks with structure tags (\glsentryplural{HAN})}
}

\newsavebox{\largestimage}

\usepackage{xcolor}
\usepackage{placeins}
\usepackage{subcaption}
\usepackage{caption}
\usepackage{adjustbox}
\usepackage{enumitem}
\setlist[enumerate]{itemsep=0mm, topsep=3pt}
\setlist[itemize]{itemsep=0mm, topsep=3pt}
\newcommand{\PARA}[1]{\vspace*{0.5em}\noindent{\bf #1}}

\newcommand\EXCLUDEEMNLPDRAFT[1]{}

\aclfinalcopy{}

\title{Structure-Tags Improve Text Classification for Scholarly Document Quality Prediction}

\newcommand\scholarlyDocument{scholarly document}
\newcommand\scholarlyDocumentPlural{scholarly documents}

\author[$\dagger$]{Gideon Maillette de Buy Wenniger}
\author[$\dagger$]{Thomas van Dongen}
\author[$\ddagger$]{Eleri Aedmaa} 
\author[$\ddagger$]{\protect\\ \bf Herbert Teun Kruitbosch} 
\author[$\mathsection$]{Edwin  A. Valentijn} 
\author[$\dagger$]{\textbf{Lambert Schomaker}}

\affil[$\dagger$]{Bernoulli Institute for Mathematics,Computer Science and Artificial Intelligence \protect\\ \small University of Groningen, Groningen, The Netherlands \protect\\ \protect\small    \texttt{gemdbw AT gmail.com  \    t.a.van.dongen AT student.rug.nl} }
\affil[$\ddagger$]{Center for Information Technology, University of Groningen, Groningen, The Netherlands \protect\\ \protect\small    \texttt{e.aedmaa AT rug.nl   \    h.t.kruitbosch AT rug.nl }}
\affil[$\mathsection$]{Kapteyn Astronomical Institute, University of Groningen, Groningen, The Netherlands \protect\\ \protect\small    \texttt{ e.a.valentijn AT rug.nl} 
}

\date{}

\begin{document}

\maketitle
\begin{abstract}

Training recurrent neural networks on long texts, in particular \scholarlyDocumentPlural{},
causes problems for learning. While \acp{HAN} are effective in solving these problems, they still lose important information about the structure of the text. To tackle these problems, we propose the use of \acp{HAN} combined with \emph{structure-tags} which mark the role of sentences in the document. 
Adding tags to sentences, marking them as corresponding to title, abstract or main body text, yields improvements over the state-of-the-art for \scholarlyDocument{} quality prediction. 
The proposed system is applied to the task of accept/reject prediction on the PeerRead dataset and compared against a recent BiLSTM-based model and joint textual+visual model as well as against plain \acp{HAN}. 
Compared to plain \acp{HAN}, accuracy increases on all three domains. 
On the computation and language domain our new model works best overall, and increases accuracy 4.7\% over the best literature result.
We also obtain improvements when introducing the tags for prediction of the number of citations for 88k scientific publications that we compiled from the Allen AI S2ORC dataset. For our \ac{HAN}-system with structure-tags we reach 28.5\% explained variance, an improvement of 1.8\% over our reimplementation of the BiLSTM-based model as well as 1.0\% improvement over plain \acp{HAN}.

\glsreset{HAN}  

\end{abstract}

\section{Introduction}
Automatic prediction of the quality of scientific and other texts is a new topic within the field of deep learning. Deep learning has been successfully applied to  many \ac{NLP} problems including text classification, as well as many computer vision applications including document structure analysis. 
These successes suggest automatic quality assessment of scientific documents, while still highly ambitious, is feasible for scientific study. 

Sequential deep learning models, particularly \acp{RNN}, \acp{LSTM} and their variants, have been particularly successful for applications that require the encoding and/or generation of relatively short sequences of text, typically at most a few sentences. Applications include (short) text classification \cite{rao2016actionable}, entailment \cite{rocktschel2015reasoning} and neural \ac{MT} \cite{bahdanau2014neural, luong2015effective}.
Newer attention-based models, particularly the transformer model \cite{VaswaniEtAlTransformers2017} are even more apt at using all of the possible context when encoding sentences, further improving performance. Transformers are also used to build general sentence embeddings with the BERT model \cite{bertDevlinEtAl2018}. 
In comparison, the accurate classification of full documents remains challenging.
To be effective, a deep learning model for longer text should fulfill the following three criteria: 
\begin{enumerate}[itemsep=0mm]
 \item \emph{Trainability}: being trainable on long texts.
 \item \emph{Computational efficiency}: efficiency as well as parallelizability, to effectively use GPUs.
 \item \emph{Rich context}:  having access to rich context at sentence and document level.
 And avoiding therefore: 1) the assumption that sentences at different locations are independent,  2) the even more crippling assumption of statistical independence of document words.
\end{enumerate}

Plain sequential models such as \acp{RNN} and \acp{LSTM} model text as unstructured word sequences. This causes problems on longer texts because of the vanishing gradient  and exploding gradient problem \cite{PascanuEtAl2013}, which hampers \emph{trainability}. Gradient bounding methods including gradient clipping  \cite{Hochreiter1998VanishingGradient}, can help to reduce these problems, but provide no solution for documents with thousands of words. 
\EXCLUDEEMNLPDRAFT{Residual connections \cite{DeepResidualLearning2015, HigwayNetworks2015}, which skip over one or multiple layers,  may also increase the trainability of very deep neural networks.
But with sequential models for text, this approach does not solve the other important problem of non-parallelizability across the sequence direction 
because of the sequential dependencies. }
Transformers and BERT are not a good match for long texts either, as these models have a computational cost that grows quadratically with sentence length.
Arguably,  bag-of-word models, including models performing average pooling over word embeddings accomplish  \emph{trainability} and \emph{computational efficiency}. However, their computational cheapness is achieved at the price of making very strong statistical independence assumptions that harm prediction quality. 

A group of models does fulfill all three criteria: hierarchical versions of sequential models, in particular \acp{HAN}  \cite{han2016}. 
\acp{HAN} produce hierarchical text encodings using a hierarchical stacking of \acp{LSTM}-with attention, for the sentence and text level. This massively increases parallelization while simultaneously reducing the number of steps the gradient signal needs to be back-propagated during training, increasing learnability. \ac{HAN} text encodings can still take much context into account at every level in the representation,  thanks to the use of \acp{LSTM}. 

While \acp{HAN} are highly effective in forming adequate representations of longer texts, they are still deficient in the use of \emph{structure information} inherent in the text. The reason is simple: these models have only one ((Bi)LSTM) encoding sub-model per level in the hierarchy. This sub-model is used to encode all the inputs at that level, without access to relevant structure context. 
In this work we observe that this problem can be tackled by adding XML-like structure-tags at the beginning and end of each input sentence. 
The effectiveness of our approach is demonstrated on two tasks:
\begin{enumerate}[label=\Alph*, itemsep=-1mm]
\item Paper accept/reject prediction on the PeerRead dataset \cite{peerread2018dataset}.  
\item Number of citations prediction for \scholarlyDocumentPlural{}, on a new dataset with 88K articles compiled from the Allen AI S2ORC dataset.
\end{enumerate}
The experiments for both tasks show that using just three tags to mark abstract, title and body text, already provides substantial improvements: A) outperforming all models on the computation and language domain and \ac{HAN} without tags on all domains, B) outperforming all other models.
Larger gains can likely be made by further enriching the tag-set. 
 The proposed tagging approach is particularly useful in the domain of \scholarlyDocument{} understanding, since while these document are typically long, they are also highly structured. 

The rest of the paper is structured as follows.
In section 2 we discuss the various existing and alternative NLP models for the aforementioned tasks of quality prediction. Section 3 describes the proposed \ac{HAN} model combined with structure-tags. Section 4 and 5 respectively discuss their use for accept/reject and number of citations prediction.

\begin{figure*}[ht]

\begin{subfigure}[t]{1.0\linewidth}

 \center{
 \scalebox{0.85}{
 \includegraphics[width=\textwidth]{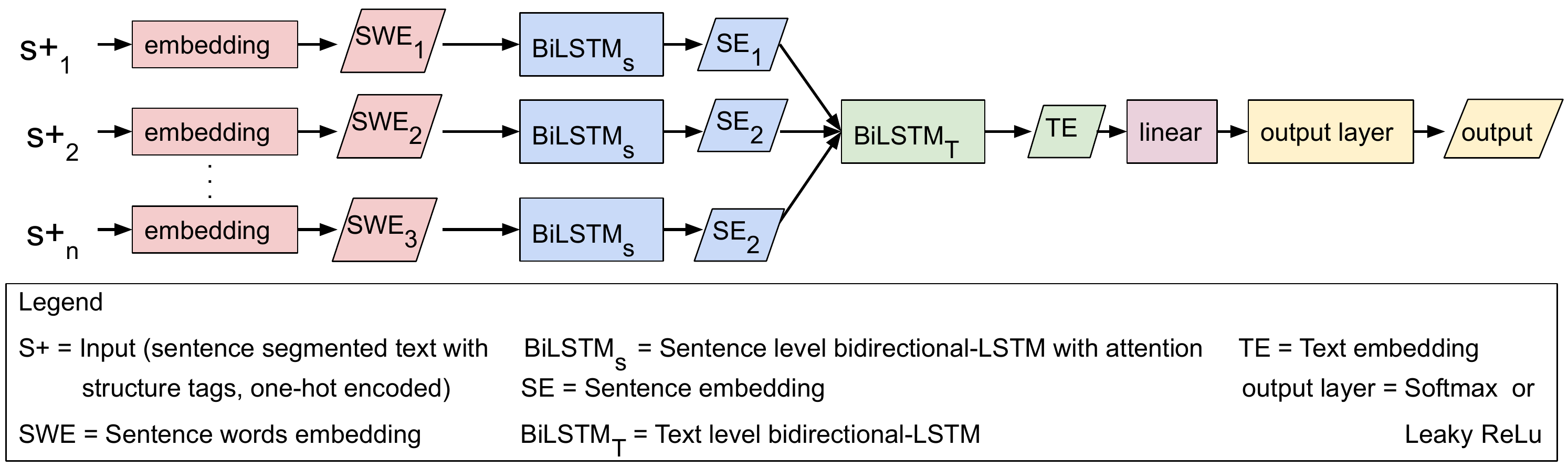}
 }}
 \caption{Our model based on \ac{HAN}.} 
\label{figure:han-model-diagram}
\end{subfigure}

\begin{subfigure}[t]{1.0\linewidth}
 \center{
 \scalebox{0.85}{
\includegraphics[width=\textwidth]{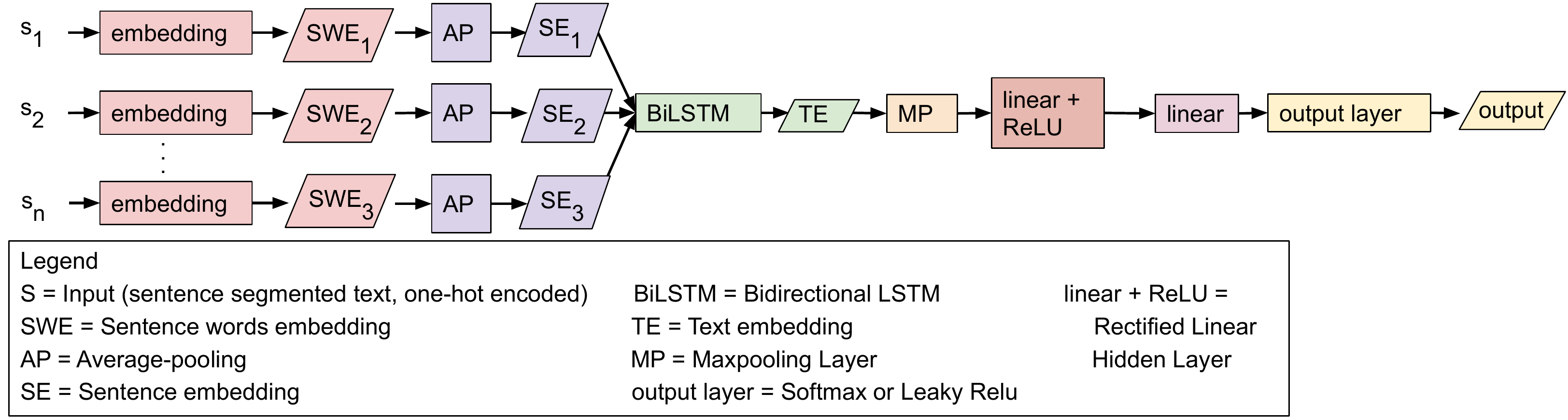}
 }
 }
 \caption{Model proposed by \citet{ShenEtAl2019}.} 
 \label{figure:shen-et-al-model-diagram}
\end{subfigure}

\caption{Most important models compared in this work.}

\end{figure*}


\section{Related Work}

Multiple methods have been proposed to estimate the quality of scientific papers. The most common approach is to use the citation counts as a measure of quality, to be predicted by models. \citet{Fu2008} proposed one of the first models which used both the papers content in the form of the paper title, abstract and keywords as well as bibliometric information. \EXCLUDEEMNLPDRAFT{such as the number of articles for the first author, publication type and quality of first author’s institution.} Notably they used automated scripts to retrieve bibliometric information, even so their final corpus is still relatively small, containing 3788 papers.
\EXCLUDEEMNLPDRAFT{
While \citet{soton260713} use information that becomes available after publication, like citation count, Fu and Aliferis use only information available before publication by using term-vectors as input to a SVM.}
\EXCLUDEEMNLPDRAFT{\citet{10.1093/bioinformatics/btp585} expands upon this research by using several different classification methods.  Both the naive Bayes as well as the logistic regression model outperform the model proposed by Fu and Aliferis.}

\EXCLUDEEMNLPDRAFT{
More recent papers use deep learning techniques to predict the citations of papers. \citet{abrishami} use recurrent neural networks to predict future citations, outperforming all other state-of-the-art methods. However, like the model proposed by Brody et al., this method is only applicable for predicting future citations when some citations are already available.}

Limited recent research is available on the subject of predicting the quality of papers with deep learning using the textual content. \citet{ShenEtAl2019} combine visual and textual content using a CNN and LSTM respectively. The authors make use of the Wikipedia and the arXiv datasets. The authors propose a joint model that classifies the quality of papers. To generate textual embeddings, the authors use a bi-directional LSTM model similar to the one proposed by the same authors in \cite{shen2017hybrid}. The input to the model is the word embeddings of a paper, obtained using GloVe, and the output is a textual embedding.

Some recent work focuses on predicting the number of citations from the paper text augmented with review text. To do so, \citet{li-etal-2019-neural} created a dataset of 
abstracts and reviews from the ICLR and NIPS conferences: 1739 abstracts with a total of 7171 reviews for ICLR and 384 abstracts with 1119 reviews for NIPS. \citet{PlankAndVanDale2019} collect a dataset of 3427 papers with 12260 reviews. Both papers show improvement in the results from using the review information.\\

\PARA{Hierarchical sequential models} \\
Hierarchical versions of sequential models have already been pioneered in the literature for a long time in the form of hierarchical \acp{RNN} \cite{HierarchicalRecurrentNeuralNetworksBengioEtAl1996}. More recently however, use of \acp{LSTM} instead of \acp{RNN} and use of attention resulted in the now popular \ac{HAN} model \cite{han2016}, which was successfully applied to sentiment analysis and text classification. \EXCLUDEEMNLPDRAFT{\citet{FengEtAl2018} used a hybrid hierarchical network, with a convolutional layer plus attention pooling layer to create article section representations and an \ac{LSTM} with attention to merge these into a final document representation. }

\PARA{Adding structure through additional inputs} \\
Our proposed structure-tag framework resembles the approach that been used for neural \ac{MT} translating multiple source languages to multiple target languages using a unified model \cite{zeroShotTranslation2016}, in which a special ``command token'' is used to indicate which kind of translation is desired. Related also is the idea of using multiple embeddings for different types of information, as introduced in neural \ac{MT} by \citet{Sennrich2016LintuisticFeatures}, which was later also exploited in the transformer model \cite{VaswaniEtAlTransformers2017}.
In contrast to the latter approaches which change the embedding layer, like \cite{zeroShotTranslation2016} we leave the (\ac{HAN}) model exactly as is and only change the input.

\section{Models}

In this work we use and refine state-of-the-art text-based deep learning models for text classification and regression tasks: accept/reject prediction and number of citations prediction respectively. Our contributions focus on \acp{HAN},\footnote{Our HAN implementation is adapted from https://github.com/cedias/Hierarchical-Sentiment} which we show for these tasks to be competitive with  models that use a flat BiLSTM encoder at their core \cite{ShenEtAl2019}. Figure \ref{figure:han-model-diagram}
shows a diagram of our \ac{HAN} model with structure-tags added to the input, and Figure \ref{figure:shen-et-al-model-diagram} shows a diagram of the BiLSTM-based model, our baseline for comparison.
As can be seen from the diagrams, both models use a BiLSTM at the text level that works on embeddings computed for the sentences of the text. 
However, while HAN uses the sequential order to compute an embedding, the baseline model averages word vectors, disregarding order, similar to bag-of-word representations.
We also use a second baseline model: Average Word Embeddings (AWE), which simply encodes text by the average word embedding.

\subsection{Sentence type tags for more structure}

The hierarchical structure of text characterized by structure elements such as sections, paragraphs and sentences and labeling elements such as document titles and section titles reveals important information. 
Models without hierarchy such as plain \ac{RNN}/\ac{LSTM} models ignore this structure, which motivated \ac{HAN}.
\ac{HAN} uses an LSTM with attention to create encodings of each sentence separately and combines this with a second LSTM with attention on top to transform these into an encoding of the entire text. The hierarchical structure of \ac{HAN} provides several advantages over flat sequential models, i.e. plain \acp{RNN}/\acp{LSTM}: 
\begin{enumerate}[topsep=2pt]
 \item \emph{Trainability on long texts}: using less steps for back-propagating gradients during training, \ac{HAN} can  process longer texts without running into vanishing/exploding gradient problems. \ac{HAN} preserves high-resolution when forming sentence-level encodings. 
 \item \emph{Computational efficiency}: 
 the structure of \ac{HAN} makes computations better parallelizable, since its sentence encoding \acp{LSTM} can process their inputs in parallel.
 \item \emph{Interpretability of predictions}: visualizing \ac{HAN} attention facilitates some qualitative insight into what inputs are important for making predictions at a sentence and word level.
\end{enumerate}

\begin{figure}
{\footnotesize
\emph{\textcolor{red}{<TITLE>}}Cross-Task Knowledge-Constrained Self Training  \emph{\textcolor{red}{</TITLE>}}  

\emph{\textcolor{red}{<ABSTRACT>}} Abstract  \emph{\textcolor{red}{</ABSTRACT>}} \\
\emph{\textcolor{red}{<ABSTRACT>}} We present an algorithmic framework for learning multiple related tasks.  \emph{\textcolor{red}{</ABSTRACT>}} \\
\ldots

\emph{\textcolor{red}{<BODY\_TEXT>}} 1 Introduction \emph{\textcolor{red}{</BODY\_TEXT>}} \\
\emph{\textcolor{red}{<BODY\_TEXT>}} When two NLP systems are run on the same data, we expect certain constraints to hold between their outputs. \emph{\textcolor{red}{</BODY\_TEXT>}} \\
\ldots

}
\normalfont
\caption{Example of structure-tags for a paper from in the PeerRead computation and language arXiv dataset.} 
\label{figure:sentence-type-tags-example}
\end{figure}

Despite these large advantages, \ac{HAN} in its normal application still remains limited in its use of structure. In particular, while \ac{HAN} encodes sentences in a hierarchical way, it does so while using the same \ac{LSTM} encoder for every sentence without structure context.
In this work we introduce a way to overcome these problems by adding sentence type tags encoding the role of a sentences or other information, which is then directly available to the BiLSTM when encoding the sentences. This is illustrated in Figure \ref{figure:sentence-type-tags-example}. First the input is segmented into a list of sentences,\footnote{We use spaCy for this: https://spacy.io/} as is also done in preprocessing for regular \ac{HAN}. Then the role of each sentence is added at the beginning and end of each sentence. In our current experiments the roles are restricted to three options: {TITLE, ABSTRACT, BODY\_TEXT}, however, the idea is general enough to include much more specific tags as well as tags encoding relative or absolute sentence position information; to be explored in future work. 
We will refer to this system as \ac{HAN-ST}.
The advantages of the tag-base approach over other possible solutions, such as using different BiLSTMs for different types of sentences are simplicity and scalability.  Equally important, using tags allows the BiLSTM to only specialize its functioning to specific types of sentences where needed, while effectively sharing what can be generalized independent of sentence type.

\begin{table*}[ht]
\caption{Data sizes and division between the ratio of accepted and rejected papers for the arXiv subsets}
\centering
\scalebox{0.80}{
 \begin{tabular}[H]{|c|c|c|c|c|c|c|c|}
 \hline
    & \multicolumn{2}{c|}{training} & \multicolumn{2}{c|}{validation} & \multicolumn{2}{c|}{testing}  &  \multirow{2}{*}{total} \\
    \cline{2-7} %
    & num & acc:rej   & num  & acc:rej   & num  &  acc:rej   & \\ 
 \hline   
 machine learning    & 4543 &  36.4\% : 63.6\%   & 252 & 36.5\% : 63.5\% &  253  & 32.0\% : 68.0\%  & 5048  \\
 \hline
 
 \multirow{1}{4cm}{computation \& language}  & \multirow{1}{*}{2374} & \multirow{1}{*}{24.3\% : 75.7\%} & \multirow{1}{*}{132}  & \multirow{1}{*}{22.0\% : 78.0\%} & \multirow{1}{*}{132}  &  \multirow{1}{*}{31.1\% : 68.9\%} & \multirow{1}{*}{2638} \\
 \hline
 artificial intelligence & 3682 & 10.5\% : 89.5\% & 205 & 8.3\% : 91.7\% & 205  & 7.8\% : 92.2\% & 4092 \\
 \hline
  
 \end{tabular}
 }
\label{table:PeerReadArxivDataSetSizes}
\end{table*}

\section{Accept/Reject prediction on PeerRead}

The first \scholarlyDocument{} quality prediction task we test our methods on is accept/reject prediction on arXiv papers from the PeerRead dataset \cite{peerread2018dataset}.
This dataset is chosen because of the large amount of earlier work in the literature reporting results on it, allowing comparison against the state-of-the-art on a well studied task.
\begin{table}
\caption{Hyperparameters used in the experiments.}
\centering
\scalebox{0.80}{
 \begin{tabular}[H]{|l|l|l|}
\hline
                          & \multirow{2}{2cm}{PeerRead classification}    & \multirow{2}{1.0cm}{S2ORC regression}       \\
& & \\                          
\hline                          
optimizer, learning rate               & \multicolumn{2}{c|}{Adam, 0.005}                        \\                    
maximum input characters &  \multicolumn{2}{c|}{20000} \\                                              
vocabulary size &  \multicolumn{2}{c|}{10000} \\   
weight initialization    &   \multicolumn{2}{c|}{} \\                                                  
              \multicolumn{1}{|r|}{general} &  \multicolumn{2}{c|}{Xavier uniform} \\                                      
              \multicolumn{1}{|r|}{lstm}    &  \multicolumn{2}{c|}{Xavier normal} \\                                     
              \multicolumn{1}{|r|}{bias}    &  \multicolumn{2}{c|}{zero}         \\                                       
              \multicolumn{1}{|r|}{word embeddings}    &  \multicolumn{2}{c|}{GloVe}         \\
              \multirow{1}{*}{loss function}            & \multirow{1}{*}{cross entropy}              & \multirow{1}{2.1cm}{MAE} 
              \\

dropout probability      & 0.5                        & 0.2                   \\
BiLSTM hidden size       & 256                        & 100                   \\
batch size               & 4                          & 64                    \\
embedding size           & 50                         & 300                   \\
\hline
  
 \end{tabular}
}

\label{table:hyperparameters}
\end{table}

\begin{table}

\caption{Total trainable parameters per model.}
\centering
\scalebox{0.80}{
 \begin{tabular}[H]{|l|l|l|l|}
 \hline
 Task & AWE & BiLSTM & HAN/HAN$_{\textrm{ST}}$ \\
 \hline
 PeerRead & 500202 & 1657222  & 3235206 \\
 \hline 
citation prediction & 3000901 & 3402801 & 3644801\\
 \hline
 \end{tabular}
 }

\end{table}

\begin{table}[ht]
\caption{The effect of the length cutoff policy on the number of words distribution.}
 \center
 \scalebox{0.77}{
 \begin{tabular}[H]{|p{4.1cm}|c|c|}
 \hline
    & \multirow{2}{2.2cm}{average words per example} & \multirow{2}{2.2cm}{median words per example} \\
    & & \\
    \hline
   20000 chars length cutoff & 3909 $\pm$ 692 & 4076 \\ 
   \hline
   360 sentences length cutoff & \textbf{5246 $\pm$ 1717} & 5514  \\
  \hline
 \end{tabular}
}
\label{table:EffectsOfLengthCutoffPolicyOnNumberOfWordsDistribution}
\end{table}

The full PeerRead dataset holds 14784 papers in total,  each of which contains implicit or explicit accept/reject labels. Furthermore, PeerRead contains different subsets of papers. The largest subset consists of arXiv papers (11778) in three computer-science sub-domains:\footnote{Based on arXiv categories within computer science, see:
https://arxiv.org/archive/cs} machine learning (cs.LG), computation and language (cs.CL), artificial intelligence (cs.AI), and has only accept/reject labels; this is the dataset that we use. A part of the papers also include reviews (3006 papers) and  a subset of the latter also contains aspect scores (586 papers).  However, of these papers with reviews, the large majority is from NIPS (2420 papers), and those papers are all accepted. As the arXiv portion is relatively larger, and accept/reject labeled, most work has focused on the task of accept/reject prediction for the papers in this set. 

Table \ref{table:PeerReadArxivDataSetSizes} shows the sizes of the different subsets of the arXiv PeerRead dataset and their respective division in number of accept and reject examples. Note that this division is imbalanced for each of the three domains, with the least imbalance for the machine learning subset and the most imbalance for the artificial intelligence subset, in which around 90\% of the examples is rejected. These imbalances in the number of examples for each of the classes make learning harder, but can be partly overcome by using strategies such as re-sampling.

\begin{table*}
\caption{PeerRead accept/reject prediction accuracy: comparison of \ac{HAN-ST} against state-of-the-art.}
\center
\scalebox{0.87}{
 \begin{tabular}[H]{|c|c|c|c|c||c|}
  \hline
   \multirow{2}{2.8cm}{arXiv sub-domain dataset} & \multirow{2}{2.2cm}{Majority class prediction} & \multirow{2}{1.8cm}{Benchmark \hspace{0.8cm} \cite{peerread2018dataset}} & 
   \multirow{2}{2.1cm}{BiLSTM  \hspace{0.8cm} \cite{ShenEtAl2019}} & \multirow{2}{2.8cm}{Joint \hspace{1.5cm} \cite{ShenEtAl2019}} &
   \multirow{2}{2.6cm}{HAN$_{\textrm{ST}}$}\\
    & & & & & \\
     & & & & &  \\
  \hline
  artificial intelligence  & 92.2\% &   92.6\% & 91.5 $\pm$ 1.03\% & \textbf{93.4 $\pm$ 1.07}\% &  89.6 $\pm$ 1.02\%\\
  \hline
  computation \& language &  68.9\% & 75.7\% & 76.2 $\pm$ 1.30\% & 77.1 $\pm$ 3.10\% & \textbf{81.8 $\pm$ 1.91}\% \\ 
  \hline
  machine learning &  68.0\% & 70.7\% & \textbf{81.1 $\pm$ 0.83}\% & 79.9 $\pm$ 2.54\% & 78.7 $\pm$ 0.69\% \\ 

  \hline
 \end{tabular}
 }
 
 \label{table:PeerReadResults}

\end{table*}

\begin{table*}[ht]
\caption{PeerRead accept/reject prediction accuracy and \emph{AUC} (area under ROC curve) scores for our models.}
\center
\scalebox{0.87}{
 \begin{tabular}[H]{|c|c|c||c|c|c|c|}
  \hline
   \multirow{3}{2cm}{arXiv sub-domain dataset} &   \multirow{3}{1cm}{metric} & \multirow{3}{1.5cm}{Majority class prediction} & 
   \multirow{3}{2.0cm}{Average Word Embeddings} &
   \multirow{3}{2.0cm}{BiLSTM  \hspace{0.8cm} (re-implemented)} &
   \multirow{3}{2.2cm}{HAN} &
   \multirow{3}{2.2cm}{HAN$_{\textrm{ST}}$}\\
    & & & & & & \\
     & & & & & & \\
   \hline
  \multirow{2}{2cm}{artificial intelligence}  & accuracy  & 92.2\%  & 74.1 $\pm$ 0.49\% & \textbf{92.4 $\pm$ 1.02}\%& 88.9 $\pm$ 1.97 \% 
   &  89.6 $\pm$ 1.02\% \\
   \cline{2-7}
   &  AUC & 0.50 & \textbf{0.793 $\pm$ 0.0143} &  0.711 $\pm$ 0.0771 & 0.625 $\pm$ 0.042   & 0.705 $\pm$ 0.055 \\
   \hline
    \multirow{2}{2cm}{computation \& language} & accuracy & 68.9\% & 73.7 $\pm$ 0.87\% & 80.1 $\pm$ 1.91\% &  80.3 $\pm$ 2.00\%  & 
    \textbf{81.8 $\pm$ 1.91}\%\\  
    \cline{2-7}
    & AUC & 0.50 & 0.740 $\pm$ 0.010 & 0.744 $\pm$ 0.056  & 0.712 $\pm$ 0.029  &   \textbf{0.745 $\pm$ 0.011} \\
   \hline
   \multirow{2}{2cm}{machine learning} & accuracy &  67.9\% & 72.9 $\pm$ 0.60\% &  \textbf{79.6 $\pm$ 3.19}\% &    76.7 $\pm$ 2.77\% &  78.7 $\pm$ 0.69\%\\  
   \cline{2-7}
   & AUC & 0.50 & 0.662 $\pm$ 0.003 & 0.743 $\pm$ 0.025 & 0.743 $\pm$ 0.019  &  \textbf{0.758 $\pm$ 0.0149} \\
   \hline
 \end{tabular}
 }
 
 \label{table:PeerReadResultsOurModels}
\end{table*}

\begin{table*}[ht]
\caption{Results of the $\textrm{HAN}_{\textrm{ST}}$ model with a reduced structure-tag set of only two tags.}
\centering
 \scalebox{0.87}{
 \begin{tabular}{|c|c|c|c|}
 \hline
 \multirow{2}{*}{\diagbox[width=2.5cm]{metric}{domain}  }& \multirow{2}{1.6cm}{artificial intelligence} & \multirow{2}{1.8cm}{computation \& language} & \multirow{2}{1.5cm}{machine learning} \\
 & & & \\
 \hline
 accuracy & 89.6 $\pm$ 1.57\% &  79.3 $\pm$ 0.14\% & 77.2 $\pm$ 1.21\% \\
 \hline
 AUC & 0.610 $\pm$ 0.067 & 0.727 $\pm$ 0.015 & 0.759 $\pm$ 0.017\\
 
 \hline
  
 \end{tabular}
}
\label{table:results-peer-read-HAN-reduced-structure-tag-set}
\end{table*}

\subsection{Experimental Setup}

In our experiments we tried to stay close to the experimental setup used by 
\cite{ShenEtAl2019}, while deviating from their settings when necessary. We used PyTorch for our code and a single GeForce RTX 2080 Ti GPU for our experiments.
Table \ref{table:hyperparameters} gives an overview of the used hyperparameters that are shared across experiments, as well as the hyperparameters that are specific to the accept/reject prediction task.
We used Adam \cite{kingma2014AdamOptimizer} as optimizer, 
and Xavier (Glorot) \cite{XavierGlorot2010} weight initialization.
We use a considerably larger learning rate of 0.005, compared to 0.0001 used by \cite{ShenEtAl2019}.\footnote{Learning rate 0.0001 gave poor results in our experiments.} On PeerRead, we use a small batch size of 4 . This is necessary for \ac{HAN} as it uses relatively much memory, 
because \EXCLUDEEMNLPDRAFT{in contrast to the \cite{ShenEtAl2019} model,} it builds rich hierarchical BiLSTM-based representations directly from the word embeddings.
We furthermore use re-sampling on the computational language and artificial intelligence subsets, as we find that without it, due to the imbalance in the label frequencies, learning fails. The re-sampling is done for each epoch, by keeping the full subset of examples with the less frequent label, but sub-sampling an equal number of random examples from the more frequent label subset. \EXCLUDEEMNLPDRAFT{In early exploratory experiments,  we also trained models with re-weighing the loss function, with weight inversely proportional to the relative class frequency.\footnote{PyTorch CrossEntropyLoss supports this natively.} However, we found that this does not fix the  problem that the model is not learning beyond predicting always the majority class, whereas re-sampling does.}
In our experiments the training of all our models  proceeds slower than the number of epochs (60) used by \citet{ShenEtAl2019} suggests. This observation holds not only for our models but also for our reimplementation of their model, and in spite of the fact that we are using a higher learning rate. We therefore used a higher number of 360 training epochs.\EXCLUDEEMNLPDRAFT{\footnote{Up to 1440 epochs were allowed for for the Average Word Embeddings model, since validation scores revealed it learns very slowly.}}
In each experiment, we used the highest accuracy score on the validation set to select the best model, using the last epoch that achieves that score in case of ties. We trained plus evaluated every model three times, to control for optimizer instability, reporting mean and standard deviation of the metric scores.

\EXCLUDEEMNLPDRAFT{\footnote{In a pre-study, we experimented with using either the first or the last best epoch in case there were multiple that tied for the best score. Generally the last best epoch model seemed to work better in case of ties, so we select that.}}

\subsubsection{Input cutoff}
Using the full text as input is in theory preferred over using only selected text, in order not to lose information prematurely.
In practice however, this is not feasible with high resolution deep learning models such as \acp{HAN}, which take input that starts at the word level. 
To save memory and computation, models may instead start out from the sentence level, using embeddings directly as inputs.  But with simple sentence embeddings, this leads to a substantial loss of input information, which may hamper performance. Even so, \cite{ShenEtAl2019} apply this strategy in a basic way by computing the average word embedding for each sentence, and using a BiLSTM model on top of that. Nevertheless, they still use a limit on the input length, by allowing only a maximum of 350 sentences. 
With \ac{HAN}, which uses more memory and computation-intense sentence-level encodings, limiting the input length is even more crucial. 
However, rather than limiting the number of sentences, we limited the maximum number of characters, set to 20000. We found that with \acp{HAN} this gives better results, even though on average it corresponds to less words.  This is explained by the distribution over the number of words per example for each of the two length cutoff policies, see Table \ref{table:EffectsOfLengthCutoffPolicyOnNumberOfWordsDistribution}. Fixing the number of sentences causes large variance in the number of words per example, likely caused by writing style differences across authors. In contrast, fixing the number of characters by definition assures a constant input length, and hence a more constant number of words (which is proportional to number of characters). 
We believe this more constant amount of information in the input aides learning.

\subsubsection{Results}

Table \ref{table:PeerReadResults} shows our best results on the PeerRead dataset, 
using \ac{HAN-ST}. The same table also shows the previous literature results of  \cite{ShenEtAl2019} and  \cite{peerread2018dataset}.
Observe that in the computation \& language domain, we gain 4.7\% accuracy over the best of the these literature models (Joint), while on the machine learning domain and artificial intelligence domain datasets we lose 2.4\%  and 3.8\% respectively in comparison to the best performing of the literature models on these domains (BiLSTM and Joint).  
In Table \ref{table:PeerReadResultsOurModels} we show the results for both our HAN models as well as for the other models. These results show a clear and consistent improvement from \ac{HAN-ST} over plain \ac{HAN}: 1.5\% accuracy for the computation \& language domain and 2.1\% for the machine learning domain and 0.7\% for the machine learning domain. Table \ref{table:PeerReadResultsOurModels} also shows results for our own re-implementation of the BiLSTM model described by \cite{ShenEtAl2019}. This useful for comparison since 
we made some changes to the experimental setup, including the use of higher learning rate and use of re-sampling.
We observe better results with our re-implementation of BiLSTM than in the original work for these datasets where re-sampling was helpful, showing its importance for imbalanced datasets.
Statistical significance was tested with the exact two-sided McNemar's test.\footnote{The scores of the 3 runs for one system are combined at example level by taking the mode/average, i.e. simple voting.} Comparing to the AWE systems: At significance level 0.05, other systems improve significantly over AWE except for HAN on the machine learning domain. At significance level 0.01 improvements over AWE are only significant on the artificial intelligence domain for BiLSTM and HAN$_{ST}$.
Other differences are not statistically significant.

While HAN$_{ST}$ is competitive with the literature models on PeerRead, it benefits from larger training data, as is available for the task of number of citations prediction.

\subsubsection{Effects of reducing the label set}

To determine the importance of different structure tags, in particular the title marking, we performed ablation experiments in which we reduced the label set. We combined the title and abstract label into one, leaving a structure-tag set with only two tags. 
Table \ref{table:results-peer-read-HAN-reduced-structure-tag-set} shows the results. As can be seen, the smaller structure-tag set reduces performance in comparison to HAN with three structure tags on all three domains. In the computation \& language domain, the model performs worse also than HAN without structure tags on both accuracy and AUC, and in the artificial intelligence domain it performs equal in terms of accuracy but worse still on AUC. In the machine learning domain 
the model also loses performance over HAN with three structure tags, but still outperforms plain HAN. The results suggest that the titles of articles contain information that is relatively important for the model to make correct classification decisions, at least for the accept/reject prediction task with the PeerRead data. We leave study into the effect of extending the structure-tag set for future work.

\begin{figure*}[ht]

\begin{subfigure}[t]{0.55\linewidth}
\begin{subfigure}[t]{1.0\linewidth}
\hspace{1.5cm}
  \lapbox[\width]{-0.3cm}{
 \scalebox{0.70}{
 \includegraphics[width=\textwidth]{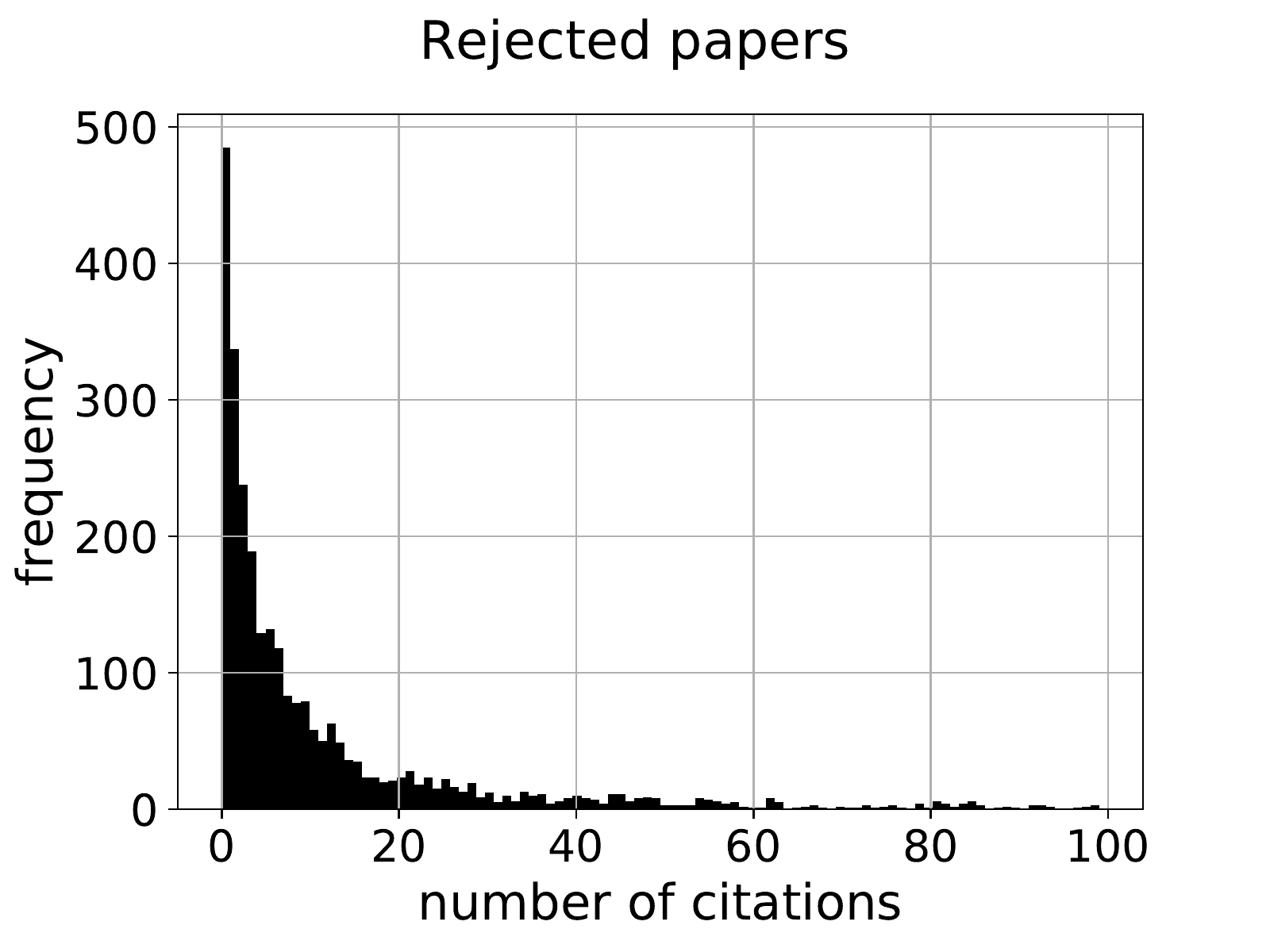}
 }
 \hspace{-0.3cm}
 \scalebox{0.70}{
 \includegraphics[width=\textwidth]{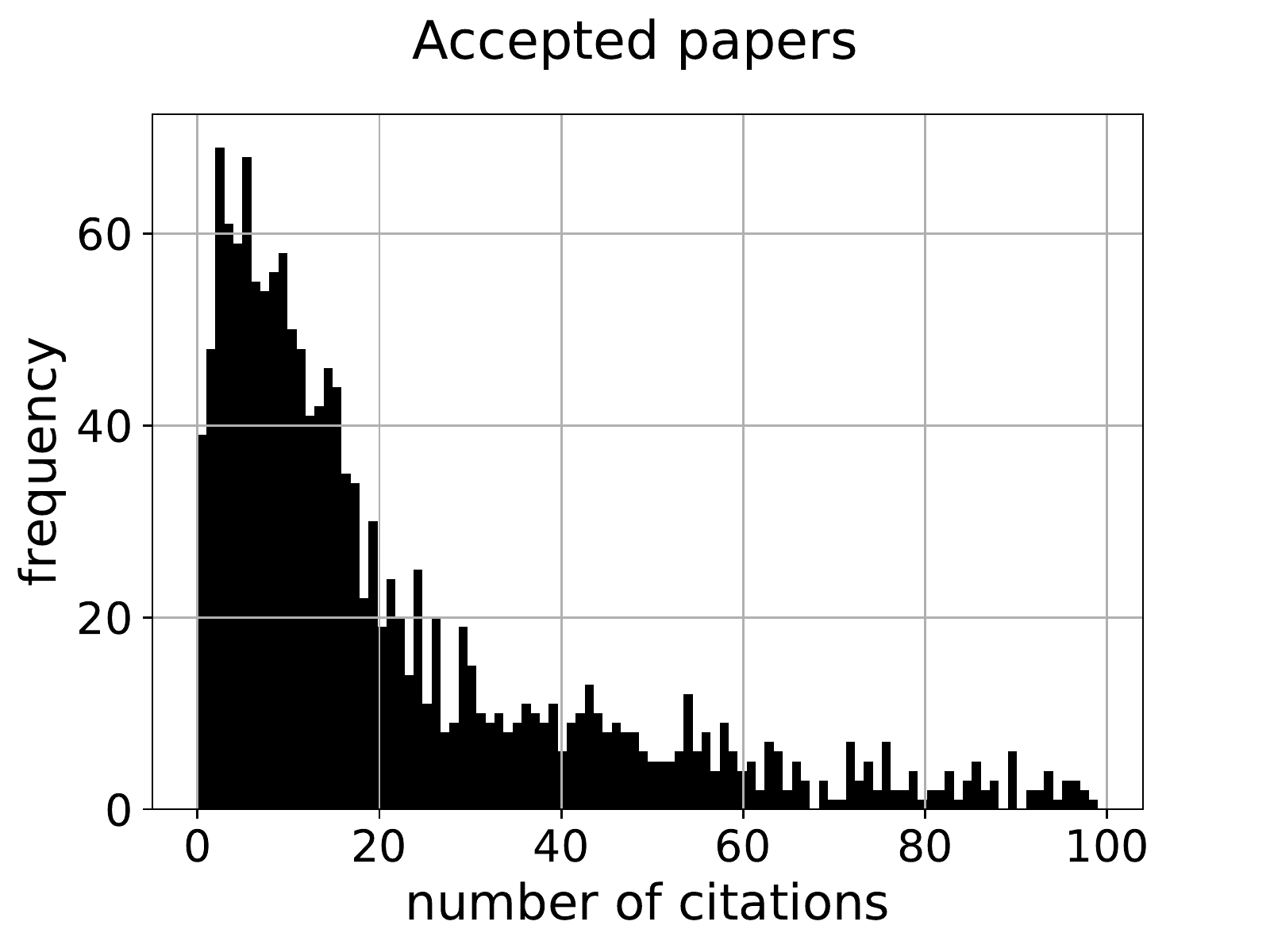}
 }
 }
 
\caption{Machine Learning domain. \hphantom{-----------}} 
 \label{figure:histograms-computation-and-language-model}
\end{subfigure}

\end{subfigure}

\begin{subfigure}[t]{0.55\linewidth}
\begin{subfigure}[t]{1.0\linewidth}
 \hspace{1.5cm}
   \lapbox[\width]{-0.3cm}{
 \scalebox{0.70}{
 \includegraphics[width=\textwidth]{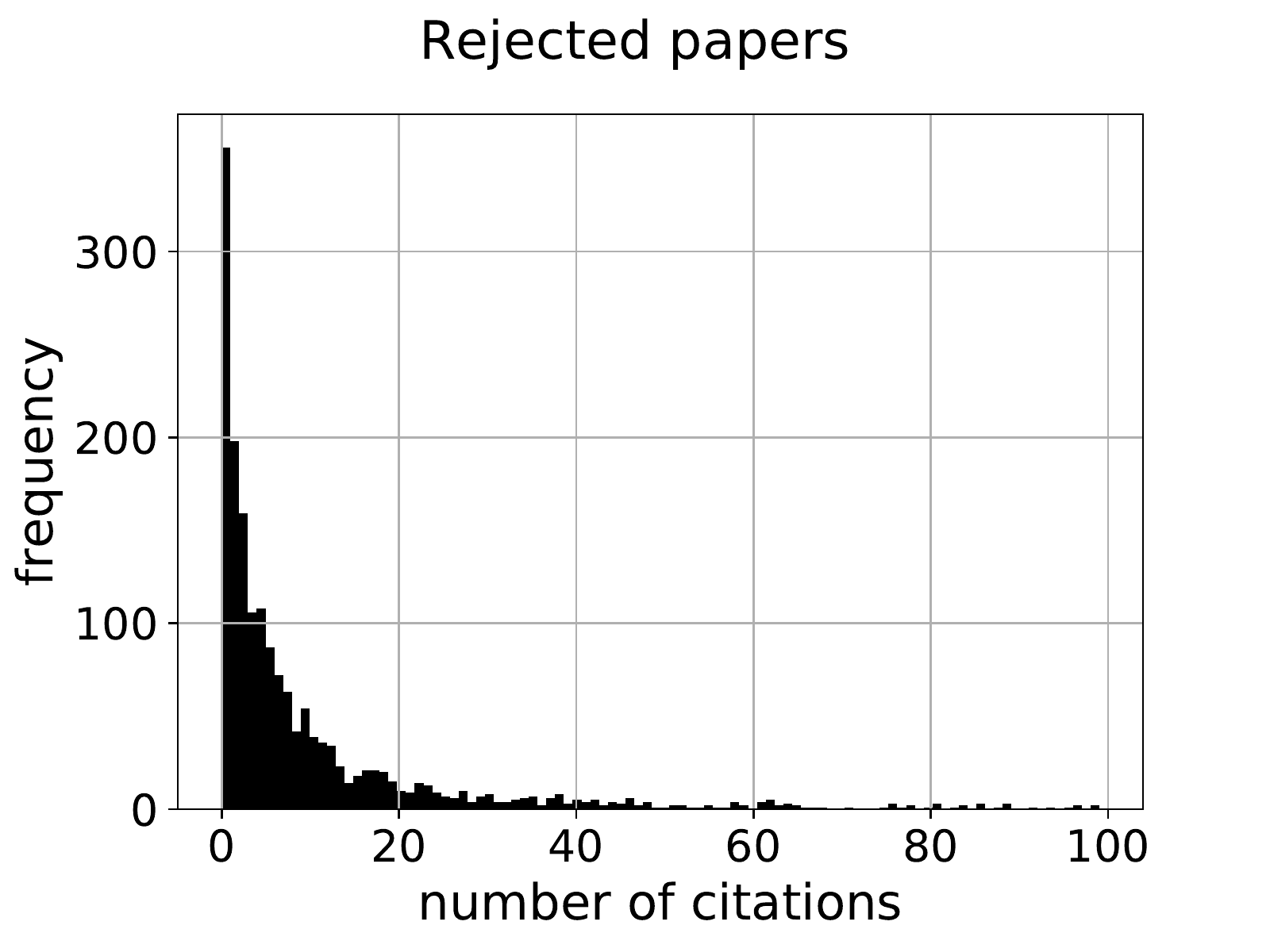}
 }
 \hspace{-0.3cm}
  \scalebox{0.70}{
 \includegraphics[width=\textwidth]{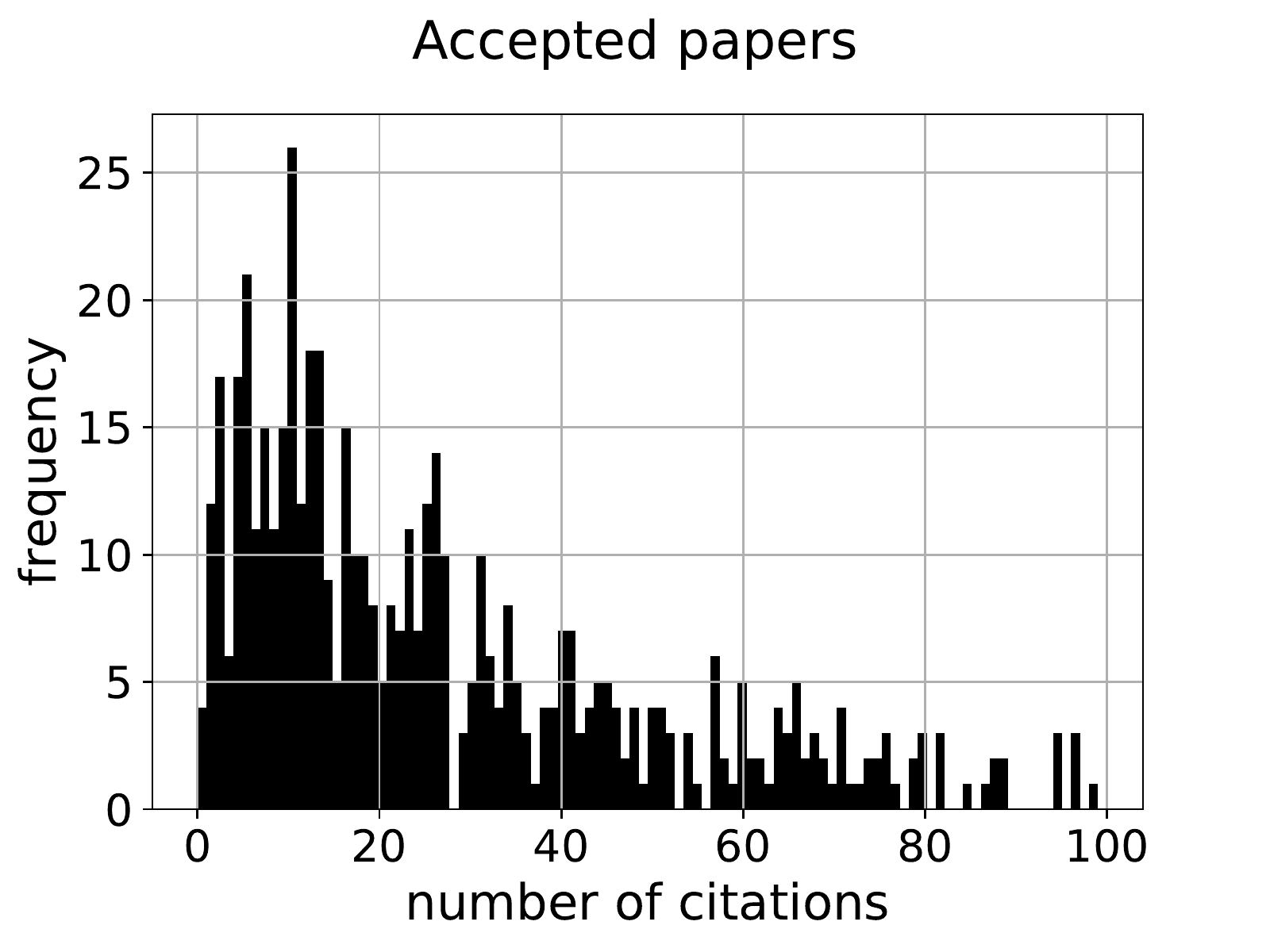}
 }
}

 \caption{Computation and Language domain.} 
 \label{figure:histograms}
\end{subfigure}
 
\end{subfigure}
%
\center
\begin{subfigure}[t]{1.0\linewidth}
  \centering
  \scalebox{0.80}{
  \begin{tabular}[H]{|c|c|c|c|}
  \hline
  \multirow{2}{*}{Domain} & \multicolumn{2}{c|}{\multirow{1}{*}{Average number of citations}} & \multirow{2}{5.5cm}{Spearman rank-order correlation coefficient ($\rho$), p-value }\\ 

    & rejected articles & accepted articles  & \\ 
   \hline     
    Machine Learning  & 24.0 $\pm$ 127.3 & 61.0 $\pm$ 232.6 & 0.375, $5\times10^{-153}$ \\
   \hline
    Computation and Language  & 14.8 $\pm$ 44.3 & 59.0 $\pm$ 105.9 &  0.466, 
   $1.6\times10^{-128}$ \\
   \hline
  \end{tabular}

}
 \caption{Global statistics.}
 \label{subtable:global-statistics}
 \end{subfigure}

\caption{Histograms and global statistics of number of citations for accepted and rejected papers for the sub-domains of PeerRead. Histograms are truncated on the right at 100 citations. 
The table in \ref{subtable:global-statistics} shows the formal correlation measure: average numbers of citations and Spearman rank-order correlation in the different domains}
\label{figure:histograms-of-peer-read-acceptance-number-of-citations-correlation}
\end{figure*}

\section{Number of citations prediction}

The second task we test our models on is number of citations prediction. A key advantage of this task is that large datasets can be obtained relatively easy by leveraging public sources such as the Semantic Scholar Database.
In contrast, obtaining accept/reject labels in large quantities typically requires having an agreement with publishers, and even then because of legal problems, it is hard to obtain and publish such data.\footnote{Note that while the PeerRead arXiv accept/reject dataset is relatively large, its labels are based on heuristics.}

Yet, how useful it is to predict the number of citations? More specifically: is the number of citations of a paper predictive of its quality? Intuitively one would expect this to be the case at least to some extent. Figure \ref{figure:histograms-of-peer-read-acceptance-number-of-citations-correlation} shows histograms of the numbers of citations of articles from the PeerRead datasets for accepted and rejected papers.\footnote{By restricting citation counting to citing papers published within two years of each paper's publication, we keep citation counts comparable across papers.} While there are some differences between the two domains, the main trend is the same in both cases: for rejected papers, the counts are peaked around zero citations and quickly decrease to one or zero for high citation counts. In contrast, the number of citations for accepted papers is two to three times higher on average, depending on the domain. Finally, we formally computed correlation in the form of the Spearman rank-order correlation coefficient ($\rho$) and associated p-value for both domains. For both domains, the value of $\rho$ is high and the p-value extremely close to zero, which indicates significant correlation can be concluded at all p-levels of significance for a two-sided test.
 These histograms and numbers prove that there is indeed a strong correlation between acceptance/rejection and the number of citations. Therefore it makes sense to consider the number of citations as an imperfect but nonetheless useful proxy for the quality of \scholarlyDocumentPlural{}.

\begin{table}
\caption{S2ORC dataset size statistics.}
\centering
\scalebox{0.87}{
 \begin{tabular}[H]{|c|c|c|}
 \hline
  data subset & num examples & avg  num words \\
  \hline
  training & 78894  & 839.1 $\pm$ 473.7 \\
  \hline
  validation & 4383   & 849.1 $\pm$ 477.5  \\
  \hline
  testing & 4382 & 856.4 $\pm$ 489.0 \\
  \hline
  
 \end{tabular}
 }
\label{table:gorc-size-statistics}
\end{table}

\begin{table*}
\caption{Test scores for the log number of citations prediction on the S2ORC dataset.}
 \center
 \scalebox{0.88}{
 \begin{tabular}[H]{|c|c|c|c|c|}
 \hline
  &  \multirow{1}{4.2cm}{Average Word Embeddings} &   \multirow{1}{4.1cm}{BiLSTM  (re-implemented)} & \multirow{1}{2cm}{HAN \hspace{2cm}} & \multirow{1}{2cm}{HAN$_{\textrm{struct-tag}}$ \hspace{2cm}} \\
  \hline
   $R^2$ score & 0.238 $\pm$ 0.0005 & 0.267 $\pm$ 0.007 & 0.275 $\pm$ 0.008 & \textbf{0.285  $\pm$  0.002} \\
   \hline
   mean squared error & 1.261  $\pm$ 0.0008 & 1.214 $\pm$ 0.009 &  1.201 $\pm$ 0.007
   &  \textbf{1.184 $\pm$ 0.002}\\
   \hline
   mean absolute error & 0.867  $\pm$ 0.0002 & 0.842 $\pm$ 0.001 &0.833 $\pm$ 0.003 & \textbf{0.831 $\pm$ 0.001} \\
  \hline
 \end{tabular}
 }
\label{table:gorc-number-of-citations-prediction-results}
\end{table*}

\subsection{The dataset}

Recent works undertake the task of number of citations prediction based on the \scholarlyDocument{} text,  but mostly do so while using relatively small datasets. As discussed in related work, some of the recent work adds review text to the input. 
However, creating models using reviewer comments limits their practical application to after reviewing and reduces available training data.
These observations motivated us to rather aim for a relatively large dataset of $\langle\textrm{paper},\textrm{number of citations}\rangle$ pairs. We selected a subset of papers in the computer science domain from the S2ORC \cite{s2orc2020} data, for which title, abstract and body text information is present; these are combined as the example text. We did this for papers in the year range 2000--2010, and counted the number of citations of citing papers that are published within 8 years after the publication of a paper.\EXCLUDEEMNLPDRAFT{\footnote{Since exact publication date is not generally available, only citation year, this is somewhat approximate.}}
Randomly ordering the papers, from this we compiled a dataset with in total about 88K papers, and statistics as shown in Table \ref{table:gorc-size-statistics}.\footnote{The new S2ORC-derived log-citation-count prediction dataset, used in our experiments, is available from: https://github.com/gwenniger/s2orc-cc/}
Note that to the best of our knowledge, the largest number of articles used for citation prediction in earlier work is described in \cite{PlankAndVanDale2019}, we use more than 23 times the number of articles used in their experiments.
While we kept the maximum number of words per example at 20000, during our experiments we have only used the first of the list of text dictionaries for each article in S2ORC , consequently the average number of words is much lower: around 840 words per example.\footnote{Due to a misunderstanding of the S2ORC data format, which actually does contain longer text when combining all the text dictionaries, which got clarified after submission.} We leave creating examples with the full paper text for future work.
The labels added to the examples consist of a function of the number of citations, as explained next.

\subsection{Citation-score as a quality proxy}
The number of citations follow of \scholarlyDocumentPlural{} follow a Zipfian distribution \cite{Silagadze1997}. That is, most papers have little citations, but those that obtain more citations tend to get exponentially more. To account for this, we used the log of the number of citations to create a metric that aims to approximates a measure of quality on a linear scale.  
In practice, we use the function: 
\setlength{\belowdisplayskip}{2pt} \setlength{\belowdisplayshortskip}{0pt} \setlength{\abovedisplayskip}{2pt} \setlength{\abovedisplayshortskip}{0pt}
\begin{equation}
\textrm{citation-score} = log_e(n + 1) 
\label{equation:citation-score}
\end{equation}
adding one to the number of citations $n$ before taking the log, to make sure the function is well-defined even for papers with zero citations. \\

\noindent \textbf{Comparison to alternative citation scores} \\
What alternatives to our log-based metric have been explored in the literature? 
\citet{li-etal-2019-neural} map citation counts to the [0,1] range, presumably by simply scaling them after the paper with the maximum  and minimum number of citations in a dataset have been determined. But this approach transfers poorly to new data, since as the number of citations follows the Zipfian distribution, still higher citation counts in unseen data are likely. Furthermore, because of the Zipfian nature of the number of citations, this transformation will map the citation score of many papers to a number close to zero, drastically inflating the evaluation scores of predictions for this citation score.
A better approach is to discretize the number of citations into a fixed number of ranges.
To  predict  the  impact  of scientific  papers,  \citet{PlankAndVanDale2019}  discretize  time-normalized citation statistics into low, medium and high impact papers based on a boxplot and outlier analysis. In comparison however, our approach does not require discretization/binning, which has advantages: 1) not committing to a fixed resolution, 2) avoiding problems for papers with a number of citations on the border of two bins, 3) allowing the predicted scores to be deterministically transformed back to a number of citations.

\subsection{Loss function and evaluation metrics}

The next important question is what loss function we should optimize when training our networks to predict the chosen citation score (1). Whereas \emph{mean-squared-error} (MSE) is the default choice for regression problems, we found this loss function to perform poorly in combination with our score. In contrast, preliminary experiments showed that \emph{mean-absolute-error} (MAE) facilitates effective and relatively stable optimization, so we decided to use this. Another important question is the choice of quality metrics. MSE and MAE are standard metrics for regression evaluation, so we report those. Additionally, we report the $R^2$ score, denoting the proportion of the variance in the dependent variable that is predictable from the independent variable(s), defined as:
\begin{equation}
 R^2 = 1 - \textrm{FVU} = 1 - \frac{\textrm{MSE}(Y,Y')}{var[Y]}
\end{equation}
With  $Y'$ and $Y$ being the predicted and actual labels respectively, MSE being the 
mean-squared-error and and FVU the fraction of variance unexplained.
This explains how the $R^2$ score normalizes for the relative difficulty for the task, by dividing by the variance of the labels in the test set.
Another interpretation is that the $R^2$ score normalizes by the error obtained when always predicting the average of the test labels. Consequently, a $R^2$ score larger than 0 means performance better than this baseline, and below 0 means worse. This avoids the need to add scores for this baseline as reference, making the $R^2$ score more directly interpretable than MSE or MAE. As such, unlike the other metrics the $R^2$ score is also meaningfully  comparable across datasets, which typically differ in test set variance.

\subsection{Number of citations prediction results}

Table \ref{table:gorc-number-of-citations-prediction-results} shows the results of our models trained on our new S2ORC number of citations prediction dataset. We observe that the \ac{HAN-ST} model outperforms the other models.
Furthermore, the improvements of \ac{HAN-ST} over BiLSTM and AWE is statistically significant (wilcoxon signed-rank test), with p-value 0.008 in both cases .

\section{Conclusion}
This work showed the usefulness of \ac{HAN} and rich context tags to the processing of scientific documents. 
Consistent improvements in prediction quality were obtained for both accept/reject estimation and number of citations prediction for \ac{HAN} when adding structure-tags. A strong and significant correlation between accept/reject labels and number of citations was demonstrated, signaling the usefulness of the latter as a measure of scholarly document quality. With more training data, as available on the citation-score prediction task, \ac{HAN} with structure-tags outperforms the strong and recently proposed scholarly document quality prediction models that we compared to in this study.

\section*{Acknowledgments}
\noindent
\includegraphics[width=1cm]{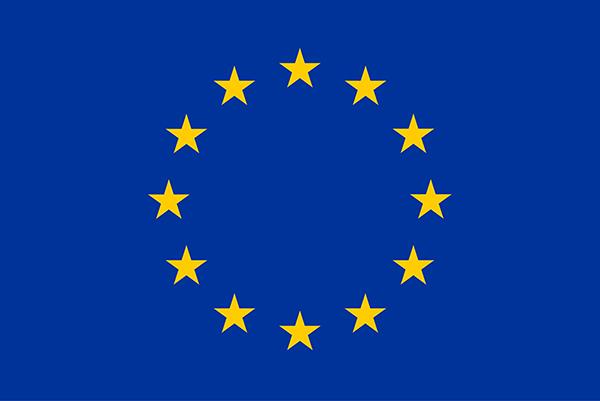}
This project has been supported by the European Fund for Regional development (EFRO) and the Target Fieldlab. The Peregrine high performance computing cluster, at the Center for Information Technology of the University (CIT) of Groningen, was used for running part of the experiments in this study.
We would like to thank the people  at the CIT for their support and access to the cluster. We would like to thank the anonymous reviewers for their helpful comments. We would also like to thank Charles-Emmanuel Dias for sharing his HAN implementation, which proved to be a solid foundation for the HAN-based models used in this work.

\bibliography{references}
\bibliographystyle{acl_natbib}

\end{document}